\documentclass[11pt]{article}

\usepackage{dialogue}

\usepackage{ifxetex}
\ifxetex
    \usepackage{fontspec}
    \setromanfont{Times New Roman}
\else
  \usepackage{cmap}
  \usepackage[T2A,T1]{fontenc}
  \usepackage[utf8]{inputenc}
  \usepackage{times}
  \usepackage{latexsym}
  \usepackage{substitutefont}
  \substitutefont{T2A}{\familydefault}{cmr}
\fi

\usepackage[russian,british]{babel}
\usepackage{url}
\usepackage{pgf}
\usepackage{makecell}
\usepackage{array}
\usepackage{multirow}

\usepackage{covington} 

\usepackage{hyperref}

\dialogfinalcopy 

\title{HAlf-MAsked Model for Named Entity Sentiment analysis}

\author{Pavel Podberezko \\
  MTS AI \\\And
  Andrey Kaznacheev \\
  MTS AI \\\And
  Sabina Abdullayeva \\
  MTS AI \\\And
  Anton Kabaev \\
  MTS AI  }

\date{}

\begin{document}
\maketitle
\begin{abstract}
Named Entity Sentiment analysis (NESA) is one of the most actively developing application domains in Natural Language Processing (NLP). Social media NESA is a significant field of opinion analysis since detecting and tracking sentiment trends in the news flow is crucial for building various analytical systems and monitoring the media image of specific people or companies. 
 
In this paper, we study different transformers-based solutions NESA in RuSentNE-23 evaluation. Despite the effectiveness of the BERT-like models, they can still struggle with certain challenges, such as overfitting, which appeared to be the main obstacle in achieving high accuracy on the RuSentNE-23 data. We present several approaches to overcome this problem, among which there is a novel technique of additional pass over given data with masked entity before making the final prediction so that we can combine logits from the model when it knows the exact entity it predicts sentiment for and when it does not. Utilizing this technique, we ensemble multiple BERT-like models trained on different subsets of data to improve overall performance. Our proposed model achieves the best result on RuSentNE-23 evaluation data and demonstrates improved consistency in entity-level sentiment analysis.

  
  \textbf{DOI:} 10.28995/2075-7182-2022-20-XX-XX
\end{abstract}

\section{Introduction}

Sentiment Analysis (SA) is a critical task in Natural Language Processing (NLP) that involves identifying the sentiment expressed in text. With the increasing amount of text data generated every day on various platforms such as social media, customer reviews, and news articles, sentiment analysis has become more important than ever. The goal of sentiment analysis is to automatically determine the emotional tone conveyed by a piece of text, which could be positive, negative, or neutral.

Over the years, several methods have been proposed for SA, ranging from traditional machine learning techniques to deep learning models based on neural networks. These methods have shown remarkable performance on different sentiment analysis tasks, such as document-level sentiment analysis, sentence-level sentiment analysis, and aspect-based sentiment analysis. In this competition, we faced one of the most significant and demanded SA problems called Named Entity Sentiment Analysis (NESA) which usually requires both identifying entities in text and determining their corresponding sentiment. However, in this case, we had to predict a sentiment label of the predetermined entity, so in this paper, we focus only on the second part. 

Despite the significant progress made in sentiment analysis, there are still several challenges needed to be addressed. For example, handling sarcasm, irony, and figurative language in the text can be challenging, as these expressions may convey a sentiment opposite to their literal meaning. In addition, named entity sentiment may come from at least three different sources: author opinion, quoted opinion, and implicit opinion.

In this paper, we review our method for Named Entity Sentiment Analysis which achieves the best result on Dialog 2023 evaluation and discuss the challenges and opportunities in this field. We also present a comprehensive overview of approaches that have been tested including all the common and uncommon competition tricks. Finally, we identify some promising directions for future research in sentiment analysis, such as developing models that can handle linguistic nuances and context-dependent sentiment.

\section{Related works}
Modern Deep Learning contains a huge domain of tasks called Text Classification Tasks like topic or intent classification, spam and fraud detection, language identification, and many others. Methodologies of solving these problems were very similar over the years and worked quite well until \cite{pang-etal-2002-thumbs} proposed to classify documents by sentiment. The authors found out that document sentiment analysis is a more challenging task to address and well-known at the time approaches were not that effective. So, researchers had to develop robust models capable of deciphering the intricate nuances of human language.

The Word2Vec model \cite{word2vec} revolutionized the field of natural language processing with its efficient training of word embeddings, which capture the semantic relationships between words. These embeddings have since become a fundamental component in many sentiment analysis models.

Recurrent Neural Networks (RNN) based approaches were dominant in the field of SA for a long time because they excel at modeling sequential data and capturing temporal dependencies in text. There are multiple improvements of this technique especially for sentiment analysis such as Recursive Neural Tensor Network \cite{RNTN} which computes compositional
vector representations for phrases of variable length
and syntactic type, CNN-BiLSTM model \cite{cnn-bilstm} which combines high-level features of document extracted by CNN and the context considered by BiLSTM that capture long-term dependency, and generalization of
the standard LSTM architecture to tree-structured
network topologies named Tree-LSTM \cite{tree-lstm}.

The introduction of the Transformer architecture \cite{transformer} marked a major breakthrough in natural language processing. With its self-attention mechanism and parallel computation capabilities, the Transformer model has become the basis for many state-of-the-art sentiment analysis systems. The appearance of such a powerful tool blurred the distinction between varieties of Text Classification tasks to some degree. Fine-tuning BERT \cite{BERT} or its any further modifications has been a crucial method of addressing this task until recently. Although, there are plenty of tricks that can vastly improve model performance in such a specific domain.
For a comprehensive understanding of this field of study, conference organizers offer an accurate and detailed observation \cite{golubev2023rusentne} of the various contributions to this domain.


\section{Methodology}
We explored and evaluated various approaches to tackle the given problem, aiming to identify the most effective technique. While we found HAlf MAsked Model (HAMAM) method to be the clear winner in terms of performance, other approaches still demonstrated notable results, deserving of honorable mention. This comprehensive assessment allowed us to not only establish the superiority of the winning technique but also gain valuable insights into the strengths and limitations of alternative methods within the context of the specific task.

\subsection{Zero-shot NESA}

\begin{figure}[htp]
  \centering
  \includegraphics[width=0.9\linewidth]{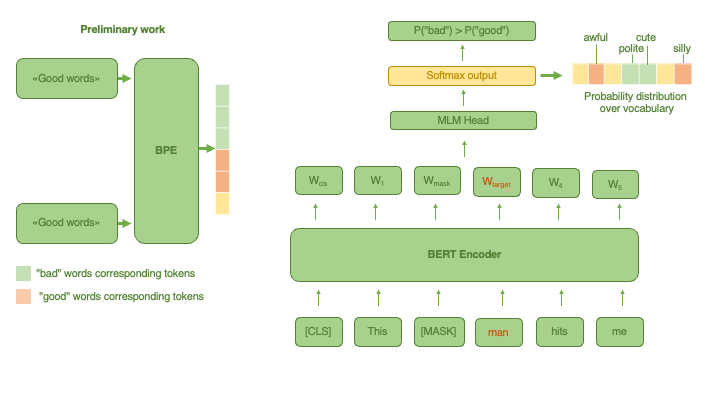}
  \caption{Scheme of zero-shot MLM-based approach.}
  \label{zeroshot}
\end{figure}

The zero-shot named entity sentiment analysis method leverages pre-trained Masked Language Models (MLMs), such as BERT, to perform sentiment classification without the need for fine-tuning or labeled data specific to the task. The following steps and figure \ref{zeroshot} provide a detailed overview of this approach:
\begin{enumerate}
    \item Insert a [MASK] token right before the target entity in the sentence. This helps the model to focus on the context surrounding the entity.

    \item Run the modified input sequence through the pre-trained MLM model, such as BERT. The model computes the probability distribution over the vocabulary for the [MASK] token based on the given context.

    \item Create two lists of tokens representing "good" and "bad" sentiment, which will be used to compute average sentiment probabilities. For each list, extract the corresponding softmax output probabilities from the model for the tokens in that list.

    \item Calculate the average probability of "good" and "bad" tokens. If the average probability of "good" tokens is higher than that of "bad" ones, classify the sentiment as positive. Otherwise, classify it as negative.
\end{enumerate}

This zero-shot method offers a straightforward approach to named entity sentiment analysis without the need for task-specific training. However, it may have limitations in handling more complex or nuanced sentiment expressions, as it relies solely on the pre-trained MLM's understanding of sentiment-related words. Moreover, it's tough to determine a neutral class because the difference between the average probabilities of "good" and "bad" tokens can be highly variable. Identifying a suitable threshold to distinguish neutral sentiment becomes difficult as the fluctuating difference makes it hard to establish clear boundaries of "approximately equal" probabilities.

\subsection{Multi-sample dropout} \label{msd}
In this section, we provide a detailed description of the multi-sample dropout technique, a regularization method presented by \cite{multisample-dropout}. This advanced approach enhances the generalization capabilities of deep learning models by employing multiple dropout masks during training for the same mini-batch of input data.

In the original dropout technique, proposed by \cite{dropout}, a single dropout mask is generated for each input instance in a mini-batch during training. This mask is applied to deactivate a random subset of neurons (or features) with a certain probability (commonly between 0.2 and 0.5), helping to prevent the model from relying too heavily on any single neuron. After applying the dropout mask, the model performs a single forward and backward pass, updating its weights accordingly.

\begin{figure}[htp]
  \centering
  \includegraphics[width=0.9\linewidth]{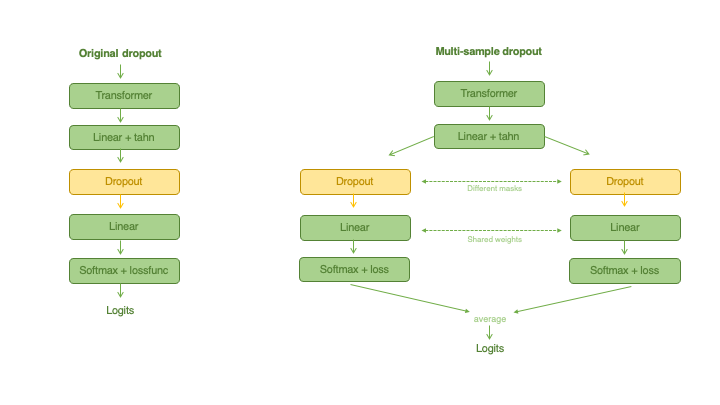}
  \caption{Overview of original dropout and multi-sample dropout.}
  \label{Multi-sample}
\end{figure}

In contrast, the multi-sample dropout applies multiple dropout masks to each input instance in a mini-batch during training. As demonstrated in figure \ref{Multi-sample}, for each input instance multiple forward passes are performed using different dropout masks, effectively exploring a broader range of neuron combinations. The outputs (logits or probabilities) from these multiple forward passes are then averaged, resembling an ensemble-like approach. The backward pass is performed using this averaged output, computing gradients and updating the model's weights. Therefore, this approach notably decreases the required number of training iterations.

In summary, while both the original dropout and multi-sample dropout techniques utilize dropout masks to improve generalization, the multi-sample dropout method extends this concept by employing multiple masks per input instance and averaging the resulting model outputs. This results in:  
\begin{itemize}
    \item accelerating training and improve generalization over the original dropout
    \item reducing a computational cost, as the majority of computation time is expended in the lower layers, while the weights in the upper layers are shared
    \item achieving lower error rates and losses
\end{itemize}

\subsection{Pooled Sentiment Model}
This method for named entity sentiment analysis combines the use of special tokens, fine-tuning, regularization techniques, and cross-validation to create a comprehensive approach that addresses overfitting and improves the model's ability to accurately predict sentiment for specific entities in a given text.

We insert a special [SENTIMENT] token before the target entity in the input text. This token serves as an indicator for the model to focus on the context surrounding the entity when predicting sentiment. Further, we chose a transformer-based model and fine-tune it to extract sentiment from the embedding corresponding to the [SENTIMENT] token. In our experiments, we tested DistilBERT (\cite{distilrubert}), BERT, and RoBERTa (\cite{roberta}). During the experiments, overfitting emerged as a challenge. To address this, various techniques were employed, including weight decay, dropout, and the utilization of weights from models trained for different tasks (such as aspect-based sentiment analysis, sentiment analysis, and Named Entity Recognition (NER)). Also, some experiments were conducted using the Monte Carlo dropout approach at inference time, which involves applying dropout during the testing phase to create an ensemble-like effect, potentially improving generalization and uncertainty estimation. 

The final model was trained using cross-validation, a technique that partitions the dataset into multiple folds, training and validating the ensemble of models on different subsets to ensure a more robust evaluation of its performance. 

This approach proved to be quite robust, and had it not been for the superior method proposed, it would have secured the 2nd position on the leaderboard.

\subsection{HAlf MAsked Model (HAMAM)}

The model builds a contextualized representation of an entity and classifies it into three given classes “positive”, “negative”, and “neutral”. As a backbone for building the representation, any transformer model can be selected. A transformer takes tokenized text as an input and produces vector representations $[h_1, ..., h_n]$ for each of the given tokens. Then two variants of entity representation are constructed:
\begin{itemize}
    \item mean pooled $v_{mean} = (h_k + ... + h_m) / (m - k)$,
    \item max pooled $v_{max} = \textrm{Max}([h_k, ..., h_m])$, with taking maximum over the last dimension,
\end{itemize}
where $k$ is the index of the first entity token and $m$ is the index of its last token.
Both $v_{mean}$ and $v_{max}$ are then passed through the classifier module, which consists of the following consecutive layers:
\begin{itemize}
    \item linear transformation $[N \times N]$, where $N$ is the size of the final hidden representation from transformer; 
    \item hyperbolic tangent function;
    \item multi-sample dropout, described in the section \ref{msd};
    \item linear transformation from $N$-dimensional vector to 3-dimensional space of target classes.
\end{itemize}

The resultant three logits are averaged for cases of mean and max pooling: $l_{entity} = (l_{mean} + l_{max})/2$. But the values of $l_{entity}$ are not used for the final prediction yet, because to avoid model overfitting to some particular words, another run of the model is performed at this point, but with the masked entity. The point is that in training data some entities might be overrepresented in one target class and underrepresented in any other, which may lead to bias in model predictions for such entities. Also while predicting any unseen entity, the model may utilize bias in the pre-trained representations of this entity. Masking the entity words (replacing them with ‘[MASK]’ token) helps to mitigate this effect and forces the model to extract sentiment information from a context rather than prior knowledge of the entity itself. 

\begin{figure}[h!]
  \centering
  \includegraphics[width=0.9\linewidth]{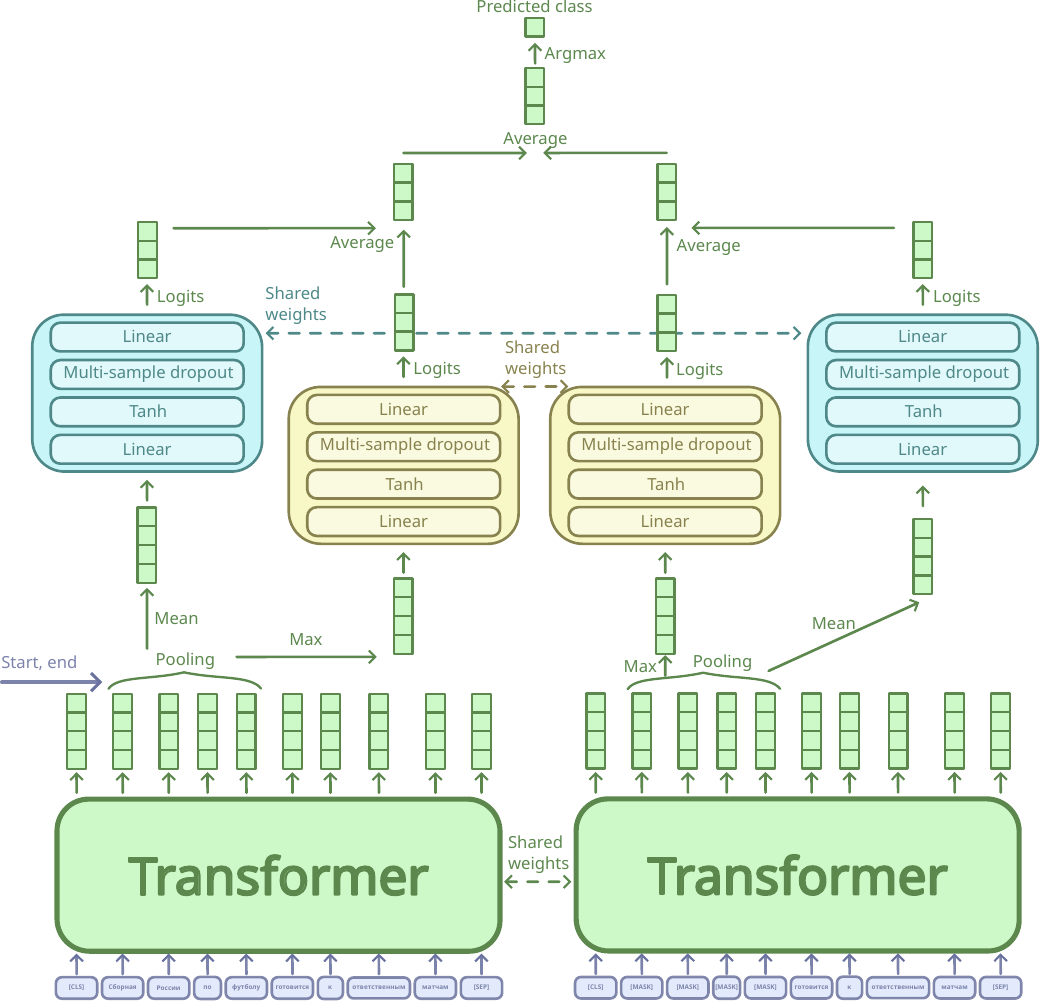}
  \caption{The architecture of HAMAM approach.}
  \label{figHAMAM}
\end{figure}

The output of the masked run is a set of logits for three target classes $l_{masked}$, which are averaged with the $l_{entity}$ before applying the argmax function to extract the predicted class. The complete architecture of the described approach is shown in figure \ref{figHAMAM}. The intuition for keeping predictions from the model with an unmasked entity and averaging it with the masked run is that despite the mentioned problems about bias, the entity itself can contain useful information for creating accurate token representations by a transformer.

During training, the loss is calculated using final logits with the weighted cross-entropy function, where weights of 1 are assigned to the examples with positive and negative sentiments, while neutral examples have weights of 0.1. The lower weight of neutral examples is motivated by, firstly, the target competition metric, which concentrates on the quality of positive and negative predictions, and also by the fact that the neutral class dominates training data as there are approximately 2.5 times more neutral examples than there are positive and negative ones combined.

Another trick motivated by the target metric and which was tested with the HAMAM approach is the threshold on neutral class prediction. Instead of taking argmax of the final logits, we first apply softmax to get probabilities of each class and in cases when the neutral class has the highest probability, but its value is below some threshold, we select the most probable class from only “positive” and “negative”.

Besides that, a well-known method for improving generalization – ensembling - was tested. In order to do that, we averaged the final logits from different transformer models trained on different subsets of training data. Specifically, the training dataset was split into five folds, each of which produces its own model trained on the four rest folds and the resultant models can be used for ensembling.

\section{Experimental setup}

Many experiments were carried out with different models and training setups. The final results for HAMAM were obtained with the following experimental setup.

The training dataset was split into 5 folds to perform cross-validation and eventually get 5 models, which can be ensembled for prediction on test data.

Training on each of the 5 parts of the initial dataset was conducted during 6 epochs with validation performed during each half-epoch. The checkpoint with the highest macro $\textrm{F}1_{\textrm{pn}}$ on validation was selected to get a score on the dev and test sets.

Several transformers were tested as a backbone for HAMAM, namely, 'DeepPavlov/distilrubert-base-cased-conversational', 'sberbank-ai/ruBert-large', 'sberbank-ai/ruRoberta-large', etc. Final results were obtained with ensemble of ‘sberbank-ai/ruRoberta-large’, ‘xlm-roberta-large’ (\cite{xlm}), and ‘google/rembert’ (\cite{rembert}).

The maximal learning rate for the backbone transformer model was set to 1e-5, while added weights for classification were trained with the maximal learning rate of 1e-4. The initial learning rate was set to 0 for all weights and warmed up to its maximal value during the first tenth part of the total training steps number and then linearly decayed to 0.

The batch size of 8 was used, and the dropout rate in classification layers was set to 0.5. In the case of multi-sample dropout, the number of samples was set to 5.

\section{Results}

Table~\ref{tabHamamConfigs} shows results for models based on the various combinations of HAMAM parts. In the most basic form, such a model performs mean pooling of the given entity and then classifies it (first line in table). In the second line, we add class weights in the cross-entropy loss calculation. The third line also adds a multi-sample dropout to this configuration, but due to the worsening of the results on local cross-validation, we removed the multi-sample dropout in the fourth line and added an entity masking trick, after which the model can be marked as HAMAM. Here we can see a large increase in cross-validation score, so we assume that entity masking is indeed helpful in avoiding overfitting and increases the model’s generalizing ability. The fifth line introduces a more sophisticated approach to entity vector pooling – a combination of ‘mean’ and ‘max’ poolings. Lines six and seven present another test of multi-sample dropout (this time model already has entity masking) both with mean and mean-max pooling correspondingly. Based on the results of cross-validation alone it is hard to tell if the last two additions (mean-max pooling and multi-sample dropout) are really helpful for the model but based on the general considerations it was decided to use the full HAMAM model for test submission. Dev scores for configurations other than full HAMAM were not obtained during the development phase of the competition, so they did not influence the model selection for the test phase.

\begin{table}[htp]
\begin{center}
\begin{tabular}{|p{0.2\textwidth}|>{\centering\arraybackslash}p{0.14\textwidth}|>{\centering\arraybackslash}p{0.14\textwidth}|>{\centering\arraybackslash}p{0.14\textwidth}|>{\centering\arraybackslash}p{0.14\textwidth}|}
\hline
Model configuration & \multicolumn{4}{c|}{Macro $\textrm{F}1_{\textrm{pn}}$} \\ \cline{2-5}
& Local cross-validation score, mean +/- std & Dev set score, mean +/- std & Dev set score from 5-fold ensemble & Test set score from 5-fold ensemble \\ \hline \hline
Mean pooling & 65.11 +/- 1.54 & 66.04 +/- 1.11 & 69.41 & 61.9 \\ \hline
\makecell[l]{Mean pooling,\\class weights in loss} & 66.11 +/- 1.05 & 66.61 +/- 0.79 & 70.47 & 65.25 \\ \hline
\makecell[l]{Mean pooling,\\class weights in loss,\\multi-sample dropout} & 65.31 +/- 1.23 & 66.72 +/- 0.66 & 70.85 & 62.84 \\ \hline
\makecell[l]{Mean pooling,\\class weights in loss,\\entity masking} & 67.83 +/- 0.45 & 67.38 +/- 0.54 & 70.86 & 65.42 \\ \hline
\makecell[l]{Mean-max pooling,\\class weights in loss,\\entity masking} & 67.63 +/- 1.07 & 67.45 +/- 0.24 & 69.14 & 65.73 \\ \hline
\makecell[l]{Mean pooling,\\class weights in loss,\\entity masking,\\multi-sample dropout} & 67.57 +/- 1.20 & 66.99 +/- 0.78 & 70.49 & 65.67 \\ \hline
\makecell[l]{Mean-max pooling,\\class weights in loss,\\entity masking,\\multi-sample dropout\\(full HAMAM)} & 67.73 +/- 1.22 & 67.20 +/- 0.31 & 69.52 & 66.25 \\ 
\hline
\end{tabular}
\end{center}
\caption{Macro $\textrm{F}1_{\textrm{pn}}$ score comparison from various configurations of HAMAM model based on 5-fold cross-validation.}
\label{tabHamamConfigs} 
\end{table}

Table~\ref{tabFinalResults} presents final results of our models on dev and test sets. HAMAM result with the threshold on the prediction of neutral class yielded a small increase in performance on the dev set, so this model configuration was used for the final submission on the test set, which gave our final test score of macro $\textrm{F}1_{\textrm{pn}}$ = 66.67.

\begin{table}[h]
\begin{center}
\begin{tabular}{|p{0.6\textwidth}|>{\centering\arraybackslash}p{0.14\textwidth}|>{\centering\arraybackslash}p{0.14\textwidth}|}
\hline
Model configuration & \multicolumn{2}{c|}{Macro $\textrm{F}1_{\textrm{pn}}$} \\ \cline{2-3}
& Dev set score & Test set score \\ \hline \hline
\makecell[l]{Pooled Sentiment model\\(5 sberbank-ai/ruRoberta-large ensemble)} & 69.92 & 65.68 \\ \hline
HAMAM (5 sberbank-ai/ruRoberta-large ensemble) & 69.52 & 66.25 \\ \hline
\makecell[l]{HAMAM (5 sberbank-ai/ruRoberta-large +\\ 4 xlm-roberta-large + 2 google/rembert ensemble)} & 70.86 & 67.0 \\ \hline
\makecell[l]{HAMAM (5 sberbank-ai/ruRoberta-large\\ + 4 xlm-roberta-large + 2 google/rembert ensemble)\\ + neutral class 0.55 threshold} & 70.94 & 66.67 \\
\hline
\end{tabular}
\end{center}
\caption{Macro $\textrm{F}1_{\textrm{pn}}$ score results from various models and their ensembles.}
\label{tabFinalResults} 
\end{table}

\section{Error Analysis}

The first thing we found when manually analyzing errors is rather ambiguous labeling. Several such examples are shown in table~\ref{tabExamples}.

\begin{table}[h]
\begin{center}
\begin{tabular}{ |m{8cm}|m{3cm}|m{1.5cm}|m{1.5cm}| } 
 \hline
 Sentence & Entity & Dataset true label & Predicted label \\ 
 \hline
 \selectlanguage{russian} На момент смерти 54-летняя журналистка расследовала коррупцию в России и нарушения прав человека в Чечне, где ранее федеральное правительство подавило попытки сепаратистов создать исламистское государство.
 & \selectlanguage{russian} правительство & negative & neutral \\ 
 \hline
 \selectlanguage{russian} 58-летний Чавес одержал в октябре победу над Каприлесом с большим численным перевесом, завоевав еще один шестилетний срок на посту президента. & \selectlanguage{russian} Чавес & negative & positive \\ 
\hline
 \selectlanguage{russian} Это был первый случай, когда сирийская армия обстреляла предполагаемых повстанцев в Ливане, который старается соблюдать нейтралитет в гражданской войне в Сирии & \selectlanguage{russian} Ливане & positive & neutral \\ 
 \hline
\end{tabular}
\end{center}
\caption{Examples of wrong predictions by HAMAM and of ambiguity in labeling.}
\label{tabExamples} 
\end{table}

Assessing human-level performance on this dataset could be intriguing.
Typically, neutral sentiment tends to be mistaken for negative and positive ones, as anticipated. Instances where the model assigns a negative sentiment to a positive label or vice versa are highly uncommon and can be attributed to ambiguous labeling. It is evident that the model has overfitted for words with highly contrasting sentiments and when they are closely associated with the entity.
For example:
{\selectlanguage{russian}“полиция задержала двоих человек возле суда и одного — внутри.”}
The model returns "negative" for {\selectlanguage{russian}“полиция”} entity.

\section{Conclusion}
In this paper, we studied different approaches for solving named entity sentiment classification task in the RuSentNE-23 competition. We presented the zero-shot technique, and also thoroughly investigated fine-tuning approach finding out that overfitting to the sentiment of certain entities is its main drawback. We described several attempts at mitigating overfitting, among which replacing entity with ‘[MASK]’ tokens showed the best result. Using this trick, we developed a new approach, which after ensembling several transformer models scored macro $\textrm{F}1_{\textrm{pn}}$ = 66.67 and reached first place in the competition.

\bibliography{dialogue.bib}

\begin{thebibliography}{}

\bibitem[\protect\citename{Chung \bgroup et al.\egroup }2021]{rembert}
Hyung~Won Chung, Thibault F{\'{e}}vry, Henry Tsai, Melvin Johnson, and
  Sebastian Ruder.
\newblock 2021.
\newblock Rethinking embedding coupling in pre-trained language models.
\newblock  // {\em 9th International Conference on Learning Representations,
  {ICLR} 2021, Virtual Event, Austria, May 3-7, 2021}. OpenReview.net.

\bibitem[\protect\citename{Conneau \bgroup et al.\egroup }2019]{xlm}
Alexis Conneau, Kartikay Khandelwal, Naman Goyal, Vishrav Chaudhary, Guillaume
  Wenzek, Francisco Guzm{\'{a}}n, Edouard Grave, Myle Ott, Luke Zettlemoyer,
  and Veselin Stoyanov.
\newblock 2019.
\newblock Unsupervised cross-lingual representation learning at scale.
\newblock {\em CoRR}, abs/1911.02116.

\bibitem[\protect\citename{Devlin \bgroup et al.\egroup }2018]{BERT}
Jacob Devlin, Ming{-}Wei Chang, Kenton Lee, and Kristina Toutanova.
\newblock 2018.
\newblock {BERT:} pre-training of deep bidirectional transformers for language
  understanding.
\newblock {\em CoRR}, abs/1810.04805.

\bibitem[\protect\citename{Golubev \bgroup et al.\egroup
  }2023]{golubev2023rusentne}
Anton Golubev, Nicolay Rusnachenko, and Natalia Loukachevitch.
\newblock 2023.
\newblock {RuSentNE-2023}: {E}valuating entity-oriented sentiment analysis on
  russian news texts.
\newblock  // {\em Computational Linguistics and Intellectual Technologies:
  papers from the Annual conference ``Dialogue''}.

\bibitem[\protect\citename{Inoue}2019]{multisample-dropout}
Hiroshi Inoue.
\newblock 2019.
\newblock Multi-sample dropout for accelerated training and better
  generalization.
\newblock {\em CoRR}, abs/1905.09788.

\bibitem[\protect\citename{Kolesnikova \bgroup et al.\egroup
  }2022]{distilrubert}
Alina Kolesnikova, Yuri Kuratov, Vasily Konovalov, and Mikhail Burtsev.
\newblock 2022.
\newblock Knowledge distillation of russian language models with reduction of
  vocabulary.

\bibitem[\protect\citename{Liu \bgroup et al.\egroup }2019]{roberta}
Yinhan Liu, Myle Ott, Naman Goyal, Jingfei Du, Mandar Joshi, Danqi Chen, Omer
  Levy, Mike Lewis, Luke Zettlemoyer, and Veselin Stoyanov.
\newblock 2019.
\newblock Roberta: {A} robustly optimized {BERT} pretraining approach.
\newblock {\em CoRR}, abs/1907.11692.

\bibitem[\protect\citename{Mikolov \bgroup et al.\egroup }2013]{word2vec}
Tom{\'{a}}s Mikolov, Kai Chen, Greg Corrado, and Jeffrey Dean.
\newblock 2013.
\newblock Efficient estimation of word representations in vector space.
\newblock  // Yoshua Bengio and Yann LeCun, {\em 1st International Conference
  on Learning Representations, {ICLR} 2013, Scottsdale, Arizona, USA, May 2-4,
  2013, Workshop Track Proceedings}.

\bibitem[\protect\citename{Pang \bgroup et al.\egroup
  }2002]{pang-etal-2002-thumbs}
Bo~Pang, Lillian Lee, and Shivakumar Vaithyanathan.
\newblock 2002.
\newblock Thumbs up? sentiment classification using machine learning
  techniques.
\newblock  // {\em Proceedings of the 2002 Conference on Empirical Methods in
  Natural Language Processing ({EMNLP} 2002)}, P 79--86. Association for
  Computational Linguistics, July.

\bibitem[\protect\citename{Socher \bgroup et al.\egroup }2013]{RNTN}
Richard Socher, Alex Perelygin, Jean Wu, Jason Chuang, Christopher~D. Manning,
  Andrew~Y. Ng, and Christopher Potts.
\newblock 2013.
\newblock Recursive deep models for semantic compositionality over a sentiment
  treebank.
\newblock  // {\em Proceedings of the 2013 Conference on Empirical Methods in
  Natural Language Processing, {EMNLP} 2013, 18-21 October 2013, Grand Hyatt
  Seattle, Seattle, Washington, USA, {A} meeting of SIGDAT, a Special Interest
  Group of the {ACL}}, P 1631--1642. {ACL}.

\bibitem[\protect\citename{Srivastava \bgroup et al.\egroup }2014]{dropout}
Nitish Srivastava, Geoffrey~E. Hinton, Alex Krizhevsky, Ilya Sutskever, and
  Ruslan Salakhutdinov.
\newblock 2014.
\newblock Dropout: a simple way to prevent neural networks from overfitting.
\newblock {\em J. Mach. Learn. Res.}, 15(1):1929--1958.

\bibitem[\protect\citename{Tai \bgroup et al.\egroup }2015]{tree-lstm}
Kai~Sheng Tai, Richard Socher, and Christopher~D. Manning.
\newblock 2015.
\newblock Improved semantic representations from tree-structured long
  short-term memory networks.
\newblock  // {\em Proceedings of the 53rd Annual Meeting of the Association
  for Computational Linguistics and the 7th International Joint Conference on
  Natural Language Processing of the Asian Federation of Natural Language
  Processing, {ACL} 2015, July 26-31, 2015, Beijing, China, Volume 1: Long
  Papers}, P 1556--1566. The Association for Computer Linguistics.

\bibitem[\protect\citename{Vaswani \bgroup et al.\egroup }2017]{transformer}
Ashish Vaswani, Noam Shazeer, Niki Parmar, Jakob Uszkoreit, Llion Jones,
  Aidan~N. Gomez, Lukasz Kaiser, and Illia Polosukhin.
\newblock 2017.
\newblock Attention is all you need.
\newblock  // Isabelle Guyon, Ulrike von Luxburg, Samy Bengio, Hanna~M.
  Wallach, Rob Fergus, S.~V.~N. Vishwanathan, and Roman Garnett, {\em Advances
  in Neural Information Processing Systems 30: Annual Conference on Neural
  Information Processing Systems 2017, December 4-9, 2017, Long Beach, CA,
  {USA}}, P 5998--6008.

\bibitem[\protect\citename{Yoon and Kim}2017]{cnn-bilstm}
Joosung Yoon and Hyeoncheol Kim.
\newblock 2017.
\newblock Multi-channel lexicon integrated cnn-bilstm models for sentiment
  analysis.
\newblock  // Lun{-}Wei Ku and Yu~Tsao, {\em Proceedings of the 29th Conference
  on Computational Linguistics and Speech Processing, {ROCLING} 2017, Taipei,
  Taiwan, November 27-28, 2017}, P 244--253. The Association for Computational
  Linguistics and Chinese Language Processing {(ACLCLP)}.

\end{thebibliography}
\bibliographystyle{dialogue}



\end{document}